\documentclass{article}
\usepackage{arxiv}

\usepackage[utf8]{inputenc} 
\usepackage[T1]{fontenc}    
\usepackage{hyperref}       
\usepackage{url}            
\usepackage{booktabs}       
\usepackage{amsfonts}       
\usepackage{nicefrac}       
\usepackage{microtype}      
\usepackage{lipsum}
\usepackage{graphicx}
\usepackage{amsmath}
\usepackage{multirow, array}
\graphicspath{ {./images/} }
\usepackage[skip=2pt,font=small]{caption}
\usepackage{atbegshi}
\AtBeginDocument{\AtBeginShipoutNext{\AtBeginShipoutDiscard}}

\usepackage{float}

\usepackage{adjustbox}

\usepackage[numbers]{natbib}
\usepackage{lipsum}
\usepackage{subcaption}
\usepackage{enumerate}
\usepackage{xcolor}

\usepackage{framed} 
\usepackage{multicol} 
\usepackage{nomencl} 
\makenomenclature

\providecommand{\keywords}[1]
{
  \small	
  \textbf{\textit{Keywords---}} #1
}

\renewcommand\nompreamble{\begin{multicols}{2}}
\renewcommand\nompostamble{\end{multicols}}

\title{Transfer learning for day-ahead load forecasting: A case study on European national electricity demand time series}

\author{
 Alexandros-Menelaos Tzortzis\\
  Decision Support Systems Laboratory
  School of Electrical and Computer Engineering\\
  National Technical University of Athens\\
  Greece\\
  \texttt{atzortzis@epu.ntua.gr} \\
    \And
 Sotiris Pelekis \\
  Decision Support Systems Laboratory
  School of Electrical and Computer Engineering\\
  National Technical University of Athens\\
  Greece\\
  \texttt{spelekis@epu.ntua.gr} \\
  \And
 Evangelos Spiliotis \\
  Forecasting and Strategy Unit
  School of Electrical and Computer Engineering\\
  National Technical University of Athens\\
  Greece\\
  \texttt{spiliotis@fsu.gr} \\
  \And
 Evangelos Karakolis \\
  Decision Support Systems Laboratory
  School of Electrical and Computer Engineering\\
  National Technical University of Athens\\
  Greece\\
  \texttt{vkarakolis@epu.ntua.gr} \\
  \And
Spiros Mouzakitis \\
  Decision Support Systems Laboratory
  School of Electrical and Computer Engineering\\
  National Technical University of Athens\\
  Greece\\
  \texttt{smouzakitis@epu.ntua.gr} \\
  \And
John Psarras \\
  Decision Support Systems Laboratory
  School of Electrical and Computer Engineering\\
  National Technical University of Athens\\
  Greece\\
  \texttt{john@epu.ntua.gr} \\
\And
Dimitris Askounis \\
  Decision Support Systems Laboratory
  School of Electrical and Computer Engineering\\
  National Technical University of Athens\\
  Greece\\
  \texttt{askous@epu.ntua.gr} \\
}

\begin{document}
\maketitle
\begin{abstract}
Short-term load forecasting (STLF) is crucial for the daily operation of power grids. However, the non-linearity, non-stationarity, and randomness characterizing electricity demand time series renders STLF a challenging task. Various forecasting approaches have been proposed for improving STLF, including neural network (NN) models which are trained using data from multiple electricity demand series that may not necessary include the target series. In the present study, we investigate the performance of this special case of STLF, called transfer learning (TL), by considering a set of 27 time series that represent the national day-ahead electricity demand of indicative European countries. We employ a popular and easy-to-implement NN model and perform a clustering analysis to identify similar patterns among the series and assist TL. In this context, two different TL approaches, with and without the clustering step, are compiled and compared against each other as well as a typical NN training setup. Our results demonstrate that TL can outperform the conventional approach, especially when clustering techniques are considered. 

\end{abstract}

\keywords{short-term load forecasting, multi-layer perceptron, national energy demand, deep learning, transfer learning, time series forecasting, ensembling}

\maketitle
\begin{table*}[!t]   
\begin{framed}
        \nomenclature{AbO}{All-but-One}
        \nomenclature{ADAM}{ADAptive Moment estimation algorithm}
        \nomenclature{CbO}{Cluster-but-One}
        \nomenclature{DL}{Deep Learning}
        \nomenclature{MAPE}{Mean Absolute Percentage Error}
        \nomenclature{ML}{Machine Learning}
        \nomenclature{MLOps}{Machine Learning Operations}
        \nomenclature{MLP}{Multi-Layer Perceptron}
        \nomenclature{NN}{Neural Network}
        \nomenclature{RMSE}{Root Mean Squared Error}
        \nomenclature{sNaive}{seasonal Naive}
        \nomenclature{STLF}{Short-Term Load Forecasting}
        \nomenclature{TL}{Transfer Learning}
        \nomenclature{TPE}{Tree-structured Parzen Estimator}
        \nomenclature{TSO}{Transmission System Operator}
\printnomenclature
\end{framed}
\end{table*}

\section{Introduction}\label{sec:introduction}
\subsection{Background}

From generation to transmission and distribution, the operation of electricity systems has to be carefully designed and planned. In this context, load forecasting is critical to properly control the operation of electricity systems and support decisions about fuel allocation, personnel management, and voltage regulation, among others. The negative implications of forecast inaccuracies in such settings have been widely explored and discussed in the literature \citep{FeinbergEugeneA.2005LoadForecasting, Alfares2002ElectricMethods, Hahn2009ElectricMaking}. Therefore, there is an increasing need for highly accurate load forecasting models that has resulted in numerous forecasting approaches through the years.  

The present study focuses on short-term load forecasting (STLF) which is most relevant for the day-to-day operation of electricity systems but also supports decisions of utilities and companies participating in electricity markets. STLF involves day- to week-ahead forecasts, while the resolution of said forecasts ranges from 15 minutes to one hour. STLF is considered a challenging task as electricity load time series are characterized by non-linearity, non-stationarity, randomness, and multiple seasonal patterns \citep{Pelekis2023ADrivers}. This is because electrical power demand originates from various electrical loads that, in turn, depend on numerous external variables, including weather and calendar factors. Additionally, the necessity for higher penetration of renewable energy sources urges for more accurate forecasts that effectively adapt to the flexibility and demand response requirements posed within transmission and distribution systems \citep{Pelekis2023TargetedTechniques}. As a result, STLF has been investigated in a variety of recent research studies \citep{Bahrami2021DeepNetworks} and projects \citep{Karakolis2022ARTIFICIALPROJECT, Wehrmeister2022TheApplications, Pelekis2023DeepTSF:Forecasting}.

Conventional STLF approaches involve the selection of a proper forecasting model and its training using historical data of the target time series (e.g. national electricity demand), often accompanied with key explanatory variables \citep{Moghaddas-Tafreshi2008a, Cui2015Short-TermModel, Haben2019a, Pelekis2022InPerformance}. Although such models are theoretically focused on a single time series, in practice, they may also be capable of accurately forecasting other series, provided that said series exhibit a sufficient level of similarity with the ones used for originally training the models. This process of "transferring" the knowledge gained from solving a certain learning problem (source task) to another, usually related problem (target task) \citep{Torrey2010TransferLearning}, is commonly referred in the literature as transfer learning (TL). The fact that TL enables the development of reusable, generalized models that can be used directly and with a minimal cost for forecasting multiple new series, has popularized the research and development on TL. This has been particularly true in applications that involve the utilization of computationally expensive models, such as deep neural networks (NNs), or the prediction of relatively small data sets. In the first case, TL can be exploited to democratize the use of sophisticated models by making them widely available to the public, while in the second case to support forecasting tasks where limited data availability would have introduced several modeling challenges. Effectively, the application of TL in the STLF domain offers the opportunity to generate accurate forecasts using the concept of pre-trained, global forecasting models \citep{Semenoglou2021InvestigatingMethods} that, with little or no retraining, can be adapted to the needs of a previously unseen (not included in the training set -- zero-shot forecasting) or slightly seen (included in the train set among many other time series -- few shot forecasting) time series.

Although the research done on TL in STLF applications has been relatively limited, the concept of TL has been extensively studied in the fields of computer vision and image recognition \citep{Chang2017ACancer, Gopalakrishnan2017DeepDetection, Kentsch2020ComputerStudy}. The main objective of TL in such settings has been the transfer of existing knowledge, representations, or patterns learned from one context and applying that knowledge to improve the performance in a different context \citep{Zhuang2021ALearning}. In contrast to traditional machine learning (ML) techniques, which require past (training) and future (testing) data to have the same domain, TL allows learning from different tasks, domains, or feature spaces. This element of TL offers immense flexibility when it comes to the source, nature, and context of the source/target data. There are various TL approaches \citep{Ribani2019ANetworks} that can be distinct based on the nature of knowledge that is being transferred, namely (a) instance transfer, (b) feature representation transfer, (c) parameter transfer, and (d) relational knowledge transfer. Depending on the approach, different techniques have 
been proposed in the literature: from simply reusing data from the source domain as part of the target domain in a process called "re-weighing" (instance transfer) \citep{Pan2010ALearning} to more complicated ones, such as "warm-start" (parameter transfer) \citep{Kunzel2018TransferNetworks, Shafahi2019AdversariallyLearning, Mitra2022AlleviatingWorkloads, Gunduz2023TransferForecasting}, "freezing/fine-tuning" (feature representation) \citep{Shafahi2019AdversariallyLearning, Mitra2022AlleviatingWorkloads},
and "head replacement" (parameter transfer) \citep{Gao2018DeepRecognition}. Note that TL techniques are not mutually exclusive. In fact, they can be blended, addressing different aspects of the forecasting problem. 


Given the benefits of TL and the limited attention it has attracted in the area of STLF, this study focuses on TL approaches for day-ahead forecasting on 27 net aggregated load time series that are available at the ENTSO-E transparency platform \citep{ENTSO-E2022ENTSO-EPlatform}. Our main objective is to investigate the potential accuracy improvements that transfer of knowledge can offer to conventional STLF NN models currently used by transmission system operators (TSOs) in Europe for forecasting  the national day-ahead electricity demand, especially within countries that share similar load patterns. This is of high interest for electrical power and energy system stakeholders as it can enable better regulating the balance between electricity production and consumption, reduce operational costs, and enhance the safety and robustness of the systems. 

We apply TL by considering an approach which combines the warm-start and fine-tuning techniques. The former technique involves copying the weights and biases from the source task model and using them to initialize the weights of the target task's models, while the latter adjusting said weights and biases on the target task. Following this approach, the time and target data needed to train the respective forecasting  models are significantly decreased, since a significant part of the solution has already been established from the source task's solution. To assess the performance of the proposed TL approach, we consider three different experimental setups. In the first setup, the forecasting models are optimized, trained, and evaluated for each country individually, i.e. in a conventional fashion and without using any TL operation, thus serving as a baseline. The second setup involves transferring information from a model trained on the 26 countries of the data set to a model tasked to generate forecasts for the respective remaining country. Finally, the third setup applies the same TL approach to the above on subsets of the data set that have been constructed using a clustering algorithm so that information is exchanged within countries of similar electricity demand patterns.

\subsection{Related work} \label{sec:1.2}


There has been abundant research on NNs and their variations on energy forecasting applications. Multi-layer perceptron (MLP) architectures are  frequently employed for solving STLF problems  \citep{Ho1992ShortAlgorithm, Kandil2006AnNetworks, Hayati2007ArtificialRegion, Arvanitidis2021EnhancedNetworks}, especially when it comes to new searching algorithms for model training \citep{Ho1992ShortAlgorithm, Mishra2008ShortOptimization, Amjady2011ASystems}. Graphical Neural Networks (GNNs) have also been utilized \citep{Wu2022EfficientNetworks} to examine the spatial correlation that data sources have with each other, while Residual Neural Networks (ResNet) models \citep{Zhang2021TransferConditions} have been proposed as an organic way for mapping complex relationships between energy loads of different regions. Feed-forward NNs have also been the topic of research concerning hyperparameter tuning and optimization \citep{Hernandez2013APlants, Farsi2021OnApproach, Lee2021IndividualizedLearning}, aiming to increase forecast accuracy and decrease computational costs of said models. Additionally, studies like those of \citet{DeFeliceMatteo2011Short-TermNotes} and \citet{VesaAndreeaValeria2020EnergyPrograms} have introduced ensembling techniques as a means of reducing parameter uncertainty in NN models, leading to more consistent load forecasts.

Focusing on deep learning (DL) NNs for STLF, recent research has put a particular emphasis on more sophisticated model architectures, such as recurrent neural networks (RNNs) and long short-term memory (LSTM) networks \citep{Zheng2017Short-TermEvaluation, Bouktif2018Optimal, KwonBo-SungandPark2020Short-TermLayer}. Many variants of RNNs and hybrid DL models \citep{Sajjad2020AForecasting,Rafi2021ANetwork} have been presented in several studies, providing forecasts concerning energy consumption at regional, national, or even building level \citep{Wang2019LSTM-basedConsumption,Memarzadeh2021Short-termAlgorithm, Yuan2021RecurrentAdaptation,Lee2022National-scaleModel}. Similarly, advanced NN architectures have gained popularity \citep{Pelekis2022InPerformance, ZhaoWenhui2022AModel, Pelekis2023ADrivers}, introducing innovative deep STLF approaches based on feed-forward \citep{Oreshkin2020N-BEATS:Forecasting, Singh2022Short-TermN-BEATS}, convolutional \citep{Yin2021Multi-temporal-spatial-scaleSystems}, or transformer \citep{Huy2022Short-TermModel, Giacomazzi2023Short-TermSources} setups.


Among the various forecasting approaches used, TL has been a key topic of research, with applications in multiple sectors. There is an extensive review done by \citet{Iman2023AAdvancements}, presenting studies involving TL in various topics including, but not limited to medical imaging, psychology, natural language processing (NLP) and quantum mechanics. TL techniques have also been employed to solve problems in the energy sector. Following the trend of aforementioned complicated architectures, models such as LSTM, ResNet, and TCN have been used to process data from multiple meters at regional \citep{Jung2020MonthlyCities}, national \citep{Lee2021IndividualizedLearning,Cai2020Two-LayerForecasting,Abdulrahman2022PredictingAttention-LSTM}, and international level \citep{Zhao2023GaussianEvents, Zhang2023GeneralCOVID-19}, 
all achieving considerable forecasting performance. Also in the context of TL, clustering techniques have been employed, aiming to identify distribution nodes that have similar trends based on the energy consumption of smart grids \citep{Syed2023InductiveGrids, Campos2023FederatedBuildings}. Ultimately, ensemble techniques have become popular in the context of TL \citep{Tan2020Ultra-Short-TermLearning, Wu2022EfficientNetworks}, once more demonstrating the importance of NN ensembles within time series forecasting applications.

Among the numerous comparative studies related to DL architectures \citep{Hu2019HeterogeneousModeling, Ma2020AData, Khalil2021TransferBuildings}, we can observe cases where MLP architectures achieve higher accuracy than other NN models \citep{Feng2020DeepData}, as well as studies demonstrating that MLPs, aided by TL, perform better than LSTM architectures \citep{Khalil2021TransferBuildings}. Another study \citep{Lee2021IndividualizedLearning} compares TL and meta-learning techniques on national data sets of Korea and Portugal, respectively, highlighting the importance of hyperparameter optimization in the process. Additionally, depending on the nature of the use case \citep{Chen2020FastNetworks, Chen2020TransferBuildings}, MLPs can be preferred over a recurrent neural network, including LSTMs. Considering these findings, alongside the numerous applications of MLP-based TL in the energy sector \citep{Demianenko2021ATechnique, Kazmi2019Large-scaleSystems, Jung2020MonthlyCities, WuDi2020MultipleForecasting}, MLPs clearly demonstrate their usefulness as a base model within the contemporary TL landscape.

\subsection{Contribution} \label{sec:1.3}

This study aims to investigate the value added by TL to STLF at an inter-country level when applied to forecast the national aggregated electricity load of different countries. In this regard, the contribution of our study is summarized as follows:


\begin{enumerate}
    \item While previous research has examined TL for STLF using national \citep{Lee2021IndividualizedLearning,Cai2020Two-LayerForecasting,Abdulrahman2022PredictingAttention-LSTM} or small-scale international transfer \citep{Zhao2023GaussianEvents, Zhang2023GeneralCOVID-19}, the present study, extends the scale of international TL to a pan-European level (27 EU countries) using load consumption data sets available at ENTSO-E \citep{ENTSO-E2022ENTSO-EWebsite}. We argue that global forecasting models can be used as a starting point to provide more accurate load forecasts by taking advantage of knowledge transfer opportunities among different, yet possibly similar electricity demand time series. This is a major contribution to the electrical power and energy system domain that can be of significant interest for TSOs and utilities, both at European and international level. 
    \item While many researchers have chosen to experiment with DL architectures \citep{Wu2022EfficientNetworks, Cai2020Two-LayerForecasting, Ho1992ShortAlgorithm, Mishra2008ShortOptimization, Amjady2011ASystems}, in this work we focus on MLPs, a relatively simple NN architecture that is well-established in the time series forecasting domain and has been proven to often outperform more sophisticated models. Apart from being faster to compute and easier to implement, MLPs allow for greater flexibility when it comes to parameter adjusting and re-training, a feature that is particularly useful in the context of TL. 
    \item Different countries may exhibit different electricity demand patterns due to key latent variables (e.g. geography, population size, and economic state). Therefore, we introduce a time series clustering approach with the objective to identify countries of similar demand patterns and facilitate TL. Although similar clustering approaches have been examined in the literature \citep{Syed2023InductiveGrids, Campos2023FederatedBuildings}, these were applied on low voltage distribution and building level data rather than country level demand time series. To this end, we introduce an additional TL setup in our study, where we apply TL among the countries of a given cluster rather than the complete set of countries available in the data set.
\end{enumerate}

\subsection{Structure of the paper} \label{sec:1.4}


The rest of the paper is structured as follows. Section \ref{sec:2} covers the methodological steps of our study including the following stages: (i) data pre-processing (sanitization of missing values, duplicates, and outliers), (ii) an exploratory data analysis that provides useful insights about the data set, (ii) the clustering process used to form groups of countries with similar load patterns, (iv) the selected model architecture, (v) the TL methodology and the respective examined setups, and (vi) the model training, validation, and evaluation procedures. Section \ref{sec:4} follows with the results derived from the investigated TL setups, followed by a brief discussion and meta-analysis of the results. Finally, in section \ref{sec:6} we conclude our work and present future perspectives.

\section{Methodology}\label{sec:2}

To effectively address our TL problem, the experimental process took place using an automated machine learning operations (MLOps) pipeline developed with MLflow \citep{Alla2021}, building up to the one described by \citet{Pelekis2022InPerformance}. The methodological steps of our approach are structured as described in the following sections.

\subsection{Data collection and curation} \label{sec:2.1}

The data set used in the present study contains the time series of the national net aggregated electricity demand of 27 different European countries \footnote{The data set includes the electricity demands of all EU countries at an hourly resolution, excluding Luxembourg and Cyprus, with the addition of Switzerland and Norway}. The data set is available online and was obtained by the ENTSO-E transparency platform \citep{ENTSO-E2022ENTSO-EPlatform}. The time series, containing a total of 1,577,954 data points, span from 2015 up to 2021, as this was the most recent version of the data collection containing full calendar years at the time when the experiments were carried out.
 
\begin{figure}[H]
    \centering
    \begin{subfigure}[b]{0.475\textwidth}
        \centering
        \includegraphics[width=\textwidth]{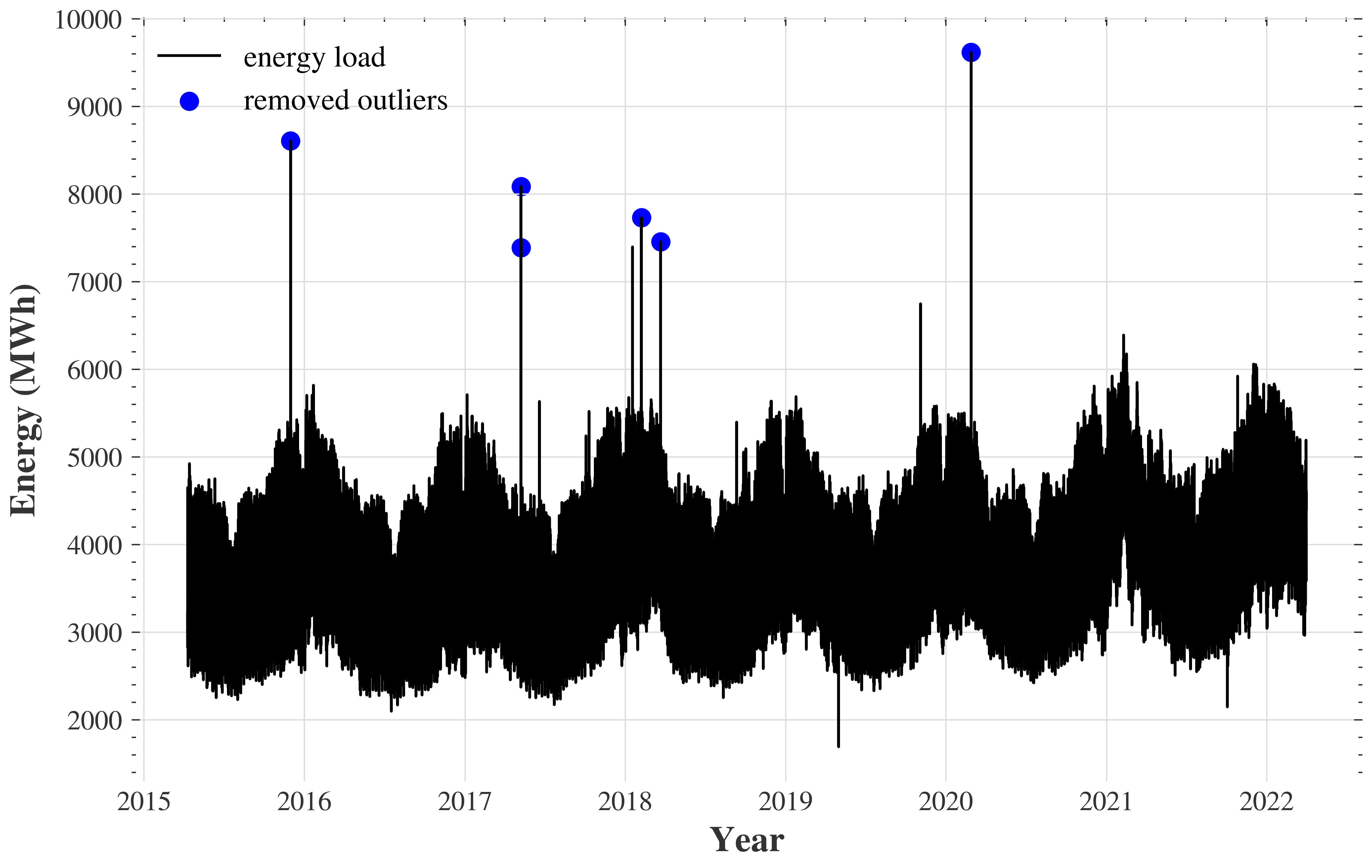}
        \caption{Extreme outliers at random data points (Denmark)}    
        \label{fig:1.1}
    \end{subfigure}
    \hfill
    \begin{subfigure}[b]{0.475\textwidth}  
        \centering 
        \includegraphics[width=\textwidth]{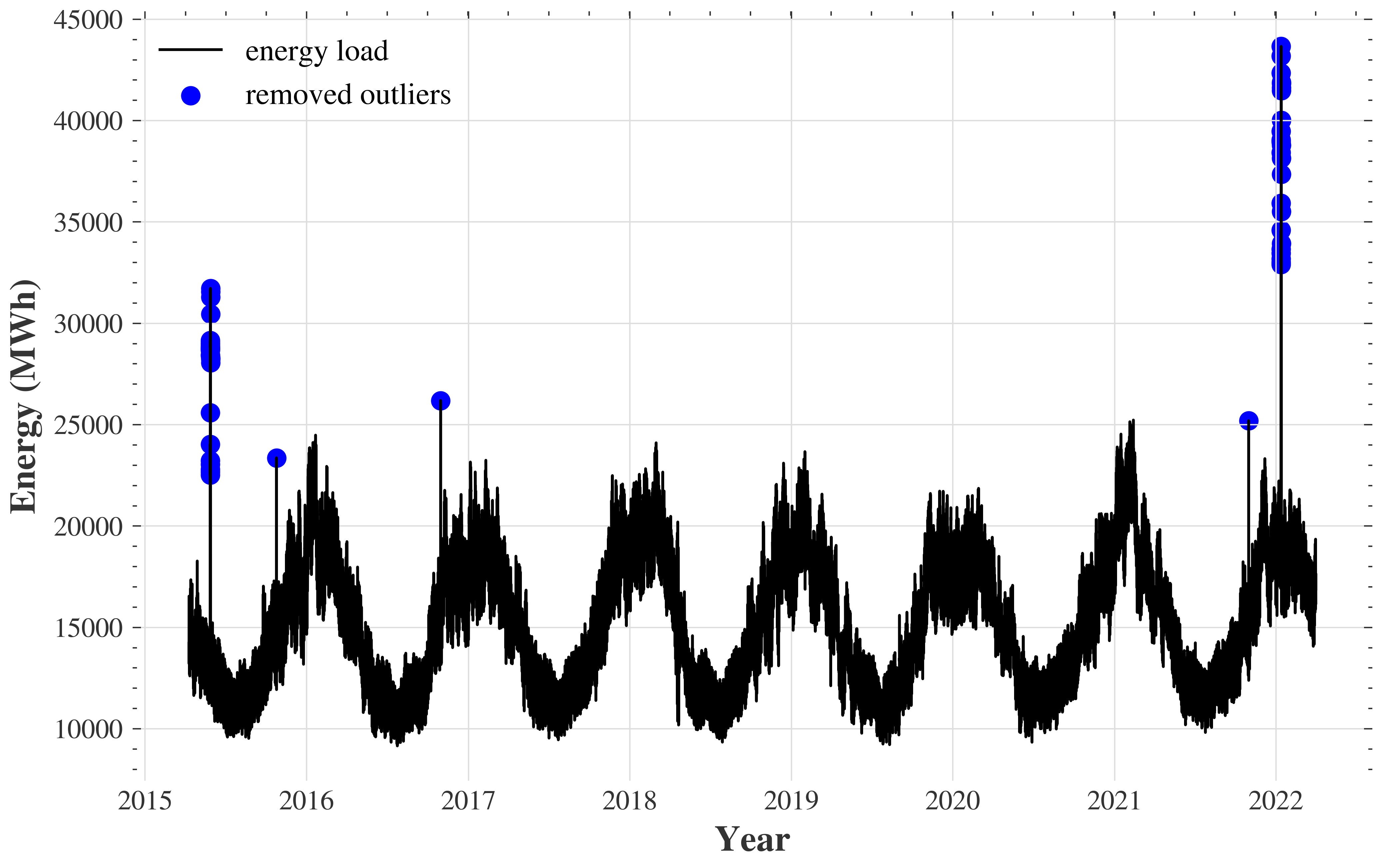}
        \caption{Outliers forming grouping at specific data points (Norway)}   
        \label{fig:1.2}
    \end{subfigure}
    \vskip\baselineskip
    \begin{subfigure}[b]{0.475\textwidth}   
        \centering 
        \includegraphics[width=\textwidth]{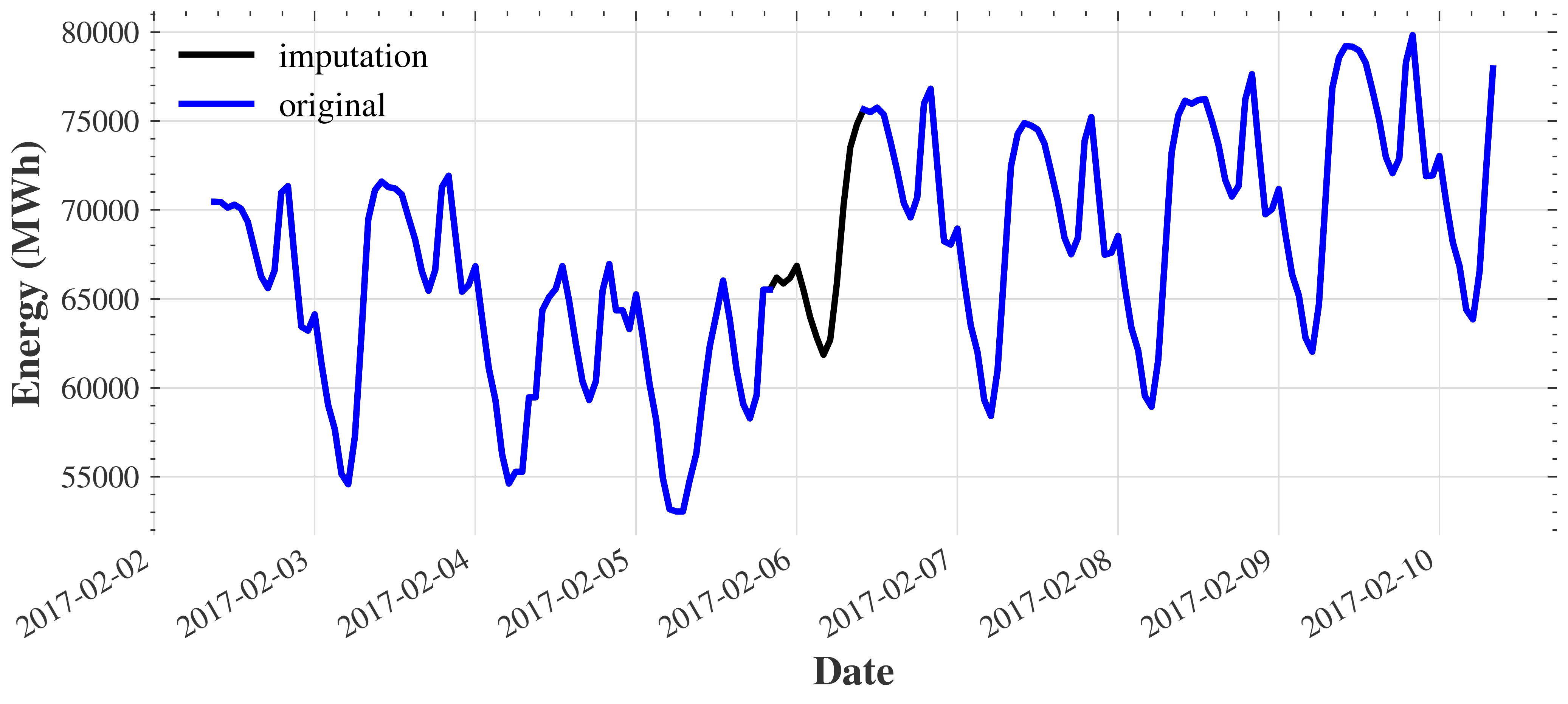}
        \caption{Imputation matching complicate pattern (France)}    
        \label{fig:1.3}
    \end{subfigure}
    \hfill
    \begin{subfigure}[b]{0.475\textwidth}   
        \centering 
        \includegraphics[width=\textwidth]{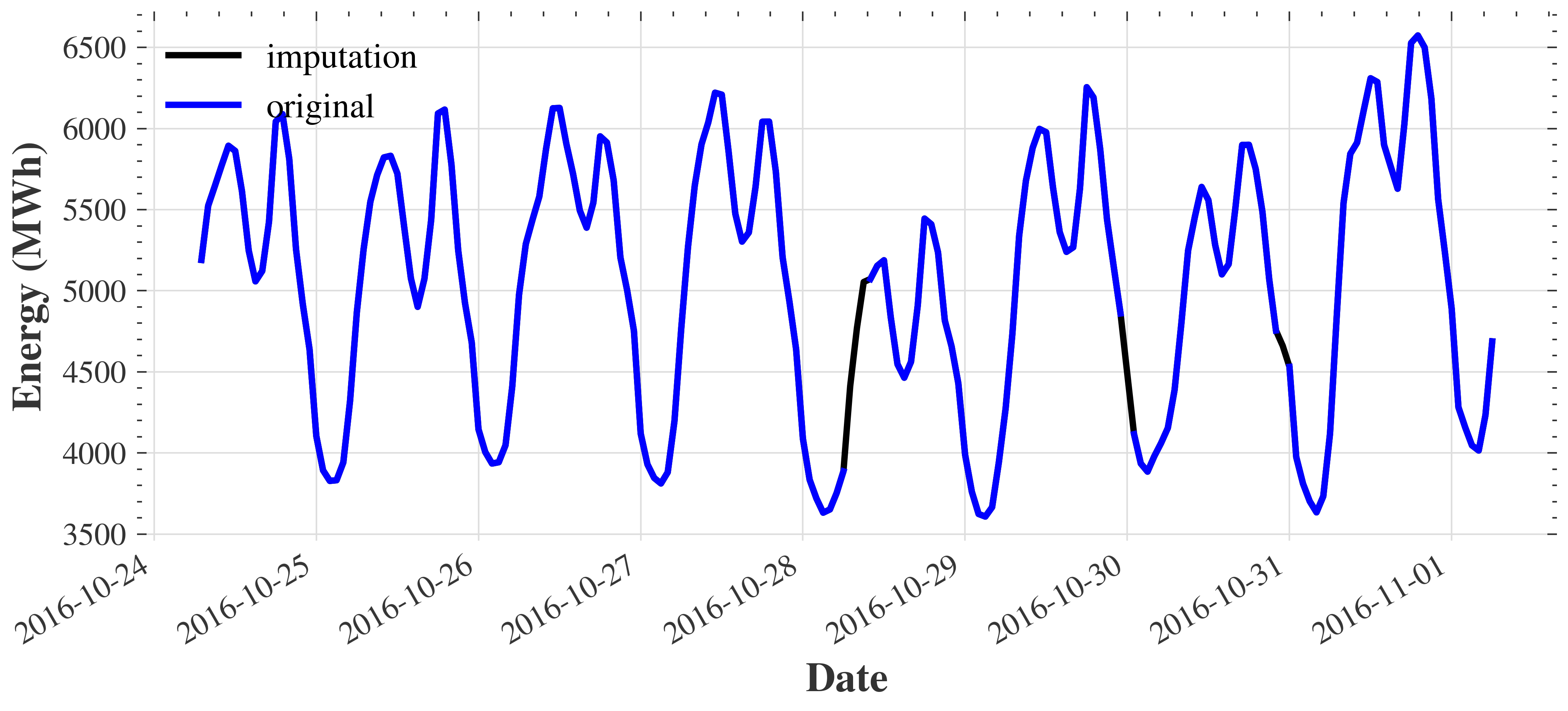}
        \caption{Imputation on multiple close proximity data points (Greece)}
        \label{fig:1.4}
    \end{subfigure}
    \caption{Visual representation of the outlier removal and missing value imputation procedures that were performed during the pre-processing of the data set.} 
    \label{fig:1}
\end{figure}

Prior to model training, data wrangling had to be carried out on the electrical load time series. To do so, we extended the methodology proposed in \citep{Pelekis2023ADrivers} to all countries included in the data set. Specifically we proceeded with the following operations:

\begin{enumerate}
    \item \textbf{removal of duplicate entries}: data may become duplicated as a result of storage or measurement errors, and therefore cleaning our data from such duplicates is necessary to avoid costly mistakes (e.g skew prediction results)  
    \item \textbf{removal of outliers}: After calculating the mean and standard deviation of the load for each month in the data set, we removed any values that arithmetically deviated from the mean more than a certain amount, aiming to exclude any outliers. More specifically, we set the maximum distance from the mean as 4.5 times the standard deviation, so as to maintain data credibility and remove only extreme outliers. This process detected 233 different outliers, with many of them forming groups in segments of our data set as shown in Fig. \ref{fig:1.1} and \ref{fig:1.2}.
    \item \textbf{conversion from native UTC to local time of each country}: Electricity demand time series can exhibit different patterns, depending on the time of the day (e.g due to daylight hours). To accommodate that and align the inter-country time series patterns through time, for each time series we inferred the timezone based on the country it referred to, changing it to local time rather than UTC which was the initial one. 
    \item \textbf{missing data imputation}: In the cases where imputation was required, we employed a hybrid method between weighted average of historical data and simple linear interpolation, similar to \citet{Peppanen2016HandlingImputation, Pelekis2023ADrivers}. The weights of each method depend exponentially on the distance of the missing value in question from the nearest timestamp that has a value according to the formula: $$w = e^{a \times d_i}, r = w \times L + (1 - w) \times H$$
    where $\alpha$ (empirically set to 0.3) is a positive weight parameter that depicts how rapidly the weight of simple interpolation drops as the distance to the closest non-empty value increases, $d_i$ is the (positive) distance (in samples) to the closest (preceding or succeeding) available sample, $L$ is the imputation suggestion by linear interpolation, $H$ is the imputation suggestion for historical data, and $r$ is the resulting imputed value. Therefore, as $d_i$ increases, the contribution of the linear interpolation suggestion decreases and the contribution of the historical data suggestion increases exponentially. Said algorithm resulted to the imputation of 13,290 data points within our data set, while an illustrative example of the process is demonstrated in Fig. \ref{fig:1.3} and \ref{fig:1.4}.
    \end{enumerate}

\subsection{Load profiling} \label{sec:2.2}
\begin{figure}[H]
    \centering
    \begin{subfigure}[b]{0.475\textwidth}
        \centering
        \includegraphics[width=\textwidth]{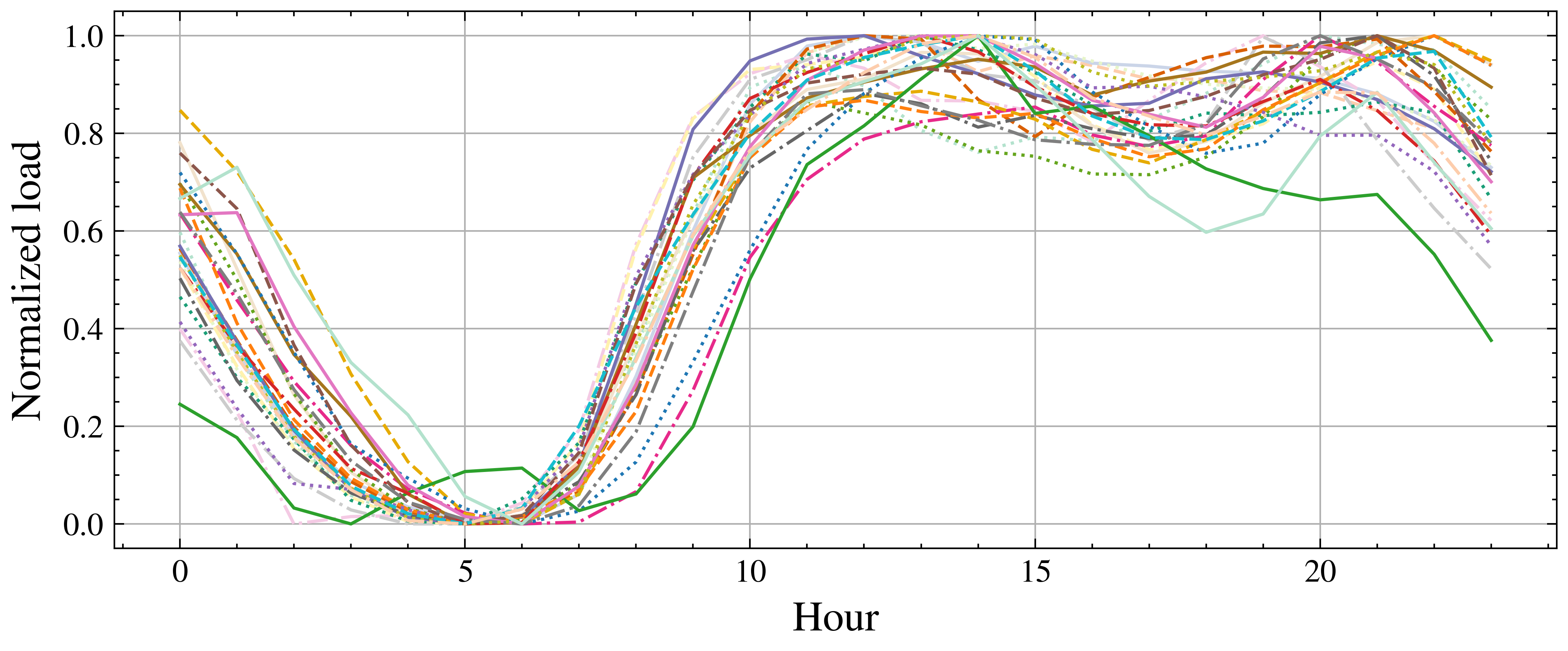}
        \caption{Average daily load profile (hourly resolution)}    
        \label{fig:2.1}
    \end{subfigure}
    \hfill
    \begin{subfigure}[b]{0.475\textwidth}  
        \centering 
        \includegraphics[width=\textwidth]{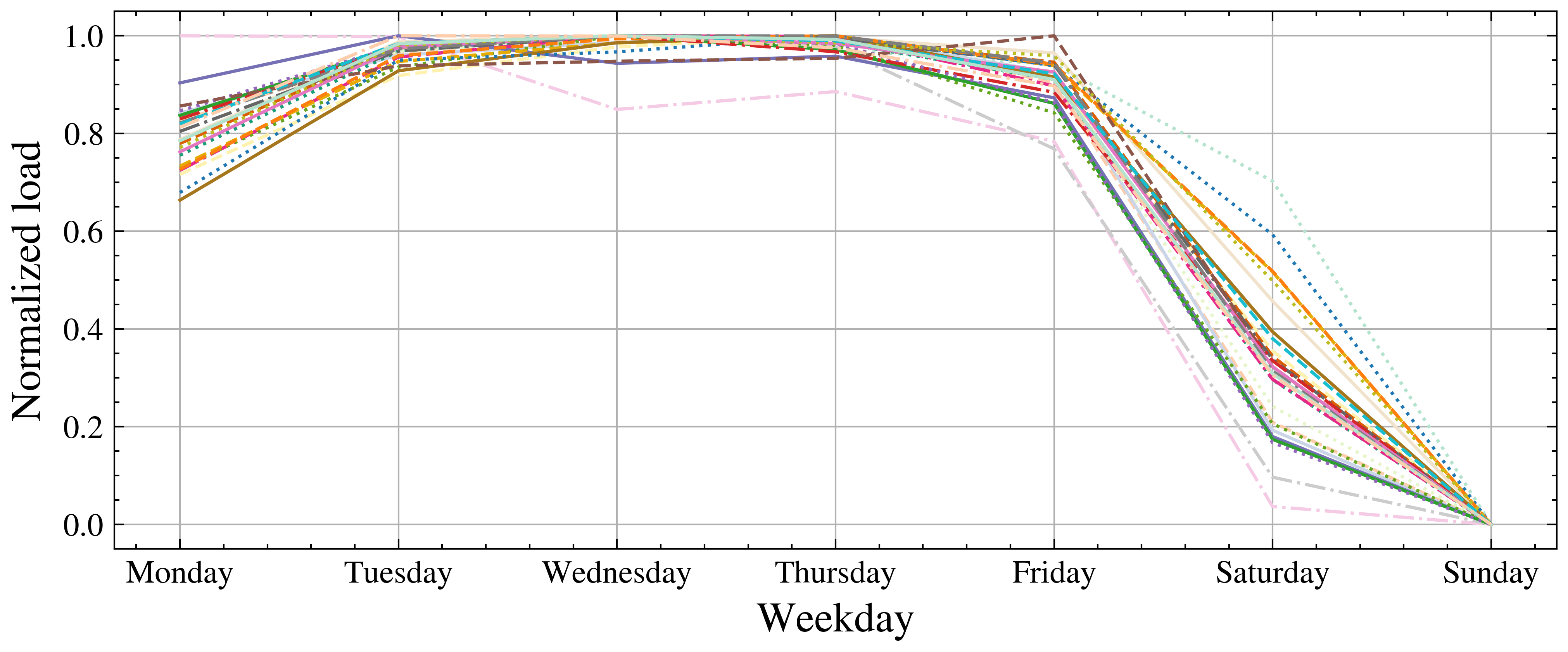}
        \caption{Average weekly load profile (daily resolution)}   
        \label{fig:2.2}
    \end{subfigure}
    \vskip\baselineskip
    \begin{subfigure}[b]{0.475\textwidth}   
        \centering 
        \includegraphics[width=\textwidth]{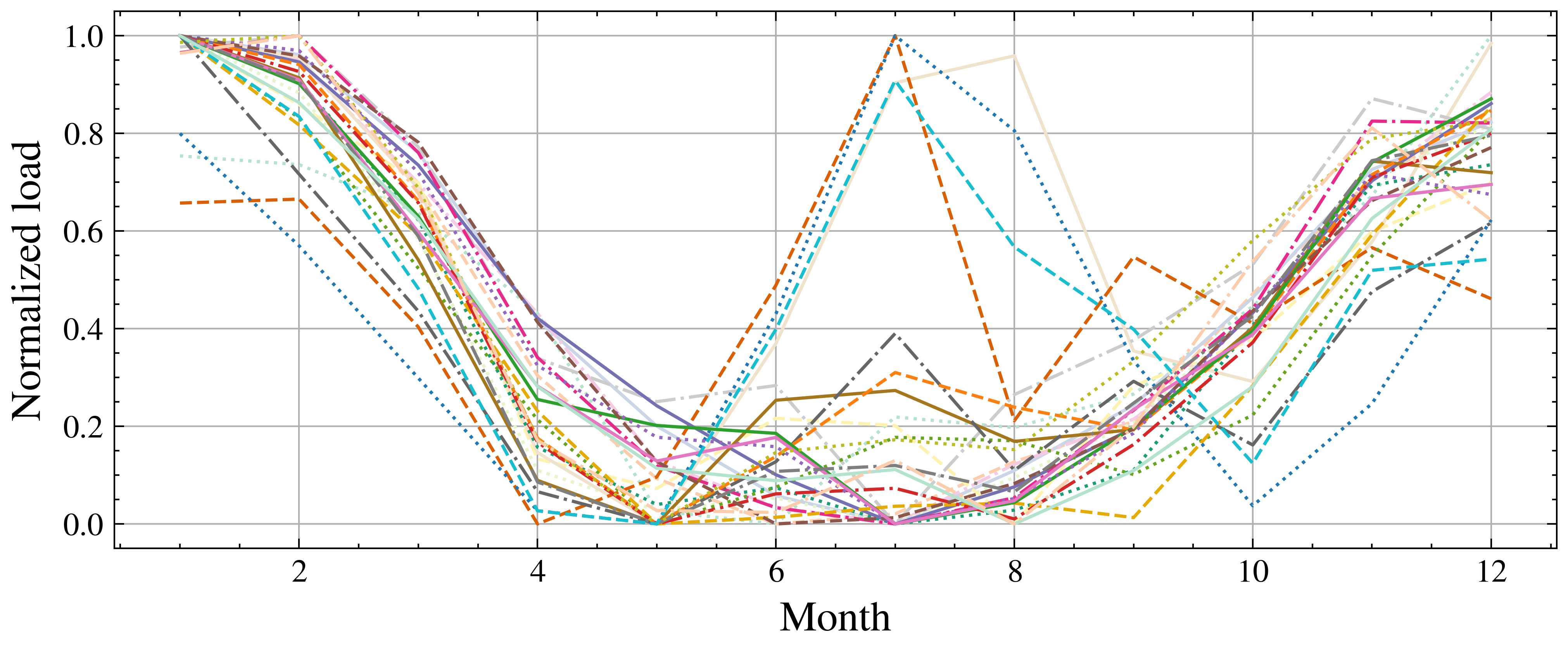}
        \caption{Average yearly load profile (monthly resolution)}    
        \label{fig:2.3}
    \end{subfigure}
    \hfill
    \begin{subfigure}[b]{0.40\textwidth}   
        \centering 
        \includegraphics[width=\textwidth]{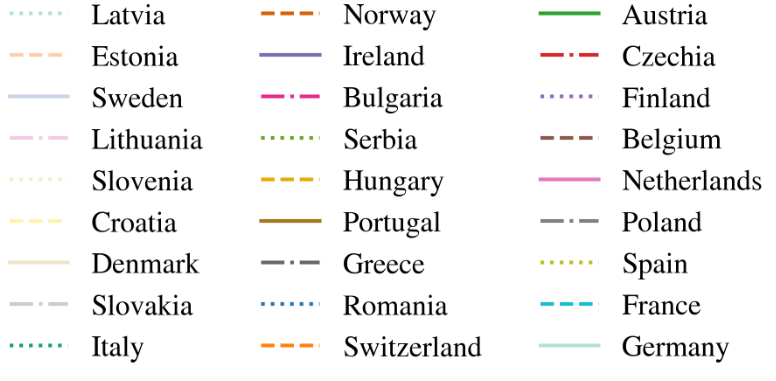}
        \label{fig:2.4}
    \end{subfigure}
    \caption{Average load profiles of the entire load time series data set aggregated by selected time intervals (hour, weekday, and month) for the time period of 2015 to 2021.} 
    \label{fig:2}
\end{figure}

Proceeding with load profiling, we initially calculated the average load profiles of the entire load time series data set, grouped by and aggregated on selected time units (hour, weekday, and month). Specifically, Fig. \ref{fig:2} illustrates the graphs of the daily (Fig.\ref{fig:2.1}), weekly (Fig.\ref{fig:2.2}) and yearly (Fig.\ref{fig:2.3}) load profiles of the countries of the data set. Subsequently, several key patterns can be noted with respect to our data, as follows:

\begin{itemize}
    \item \textbf{daily load profile:} In general, most countries exhibit a steep increase starting from sunrise, reaching a peak at noon, with a small decrease during the noon break time, followed by a steady decrease towards night hours (fig. \ref{fig:2.2}). This can be attributed to the increased energy demand during working hours, and confirmed by the decrease during noon break/lunch time where working activity is decreased. However, several countries (e.g Switzerland, France) exhibit differentiating patterns leading to the need for further investigation. 
    \item \textbf{weekly load profile:} we observe a steady energy demand during the working days and a steep decrease as we reach the weekend days (fig. \ref{fig:2.2}). This is attributed to the fact that commercial and industrial activity is decreased at the end of the week (Saturday and Sunday).
    \item \textbf{yearly load profile:} For the vast majority of countries, the summer months have on average the lowest energy requirements and they steadily increase as we move towards winter. After that, the energy demand follows the reverse pattern: steadily decreasing till it reaches the lowest again at summer (Fig. \ref{fig:2.3}). This behavior can be attributed to the increased demand of heating loads during the winter. However, the opposite trend can be observed for southern European countries with much warmer climates (Greece, Spain, Italy, Croatia, and Portugal) which demonstrate higher energy load during summer months due to the increased cooling demand.
\end{itemize} 

Taking the above into consideration, it becomes evident that most countries follow common patterns during the span of any time profile given. However, as several differences and sub-classifications can be observed -- especially within the yearly (Fig. ~\ref{fig:2.3}) and daily profiles (Fig. \ref{fig:2.1}) -- we are prompted to compare their energy loads and seek further classifications and dissimilarities through the clustering procedure of the following section.

\subsection{Clustering} \label{sec:2.3}
\begin{figure}[H]
    \centering
    \includegraphics[width=0.8\textwidth]{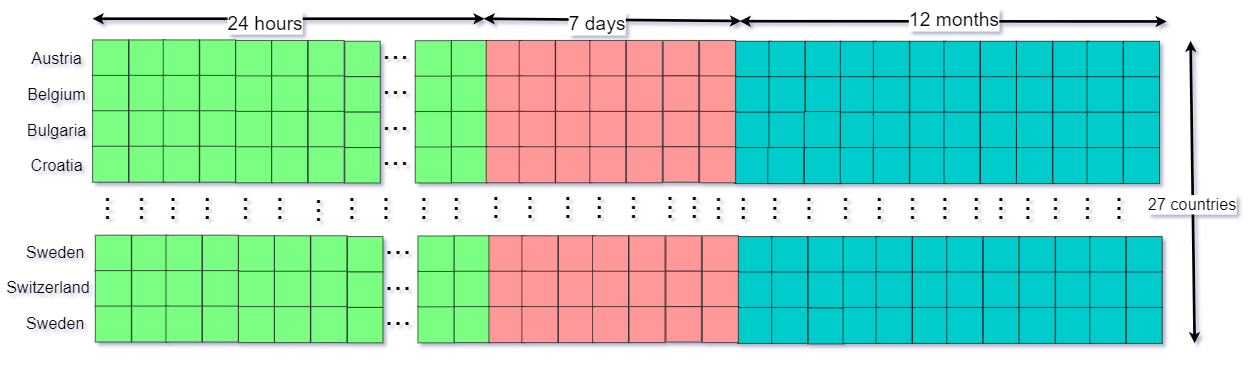}
    \caption{An illustration of the derived load profile vectors containing daily and yearly profiles for each country of the data set.}
    \label{fig:3}
\end{figure}

Motivated by the insights of the previous section regarding the inter-country load profile patterns, we are prompted towards grouping countries together in clusters that will act as source domains during TL aiming at higher forecasting performance. A hierarchical clustering approach using Ward's linkage has been identified as the best option for clustering functional data that exhibit periodic trends \citep{Ferreira2009AData,Vijaya2019ComparativeClustering}. To measure the distance between clusters (and by extension countries) our approach has been focused on the countries' timely load profiles. In this direction, each country is represented as a vector consisting of 4 sub-vectors, each one containing its daily, weekly, and yearly profiles, as depicted in~\ref{fig:3}. The vectors have been normalized to (i) ensure a clustering process of high quality that only depends on load shapes rather than magnitudes, as the latter are highly variable among countries, and (ii) enhance the computational efficiency of the algorithm. Note that longer-term load profiles have been ignored as we focus on STLF (the contribution of such profiles is negligible for the examined forecasting horizon).

\begin{figure}[H]
    \centering
    \includegraphics[width=0.75\textwidth]{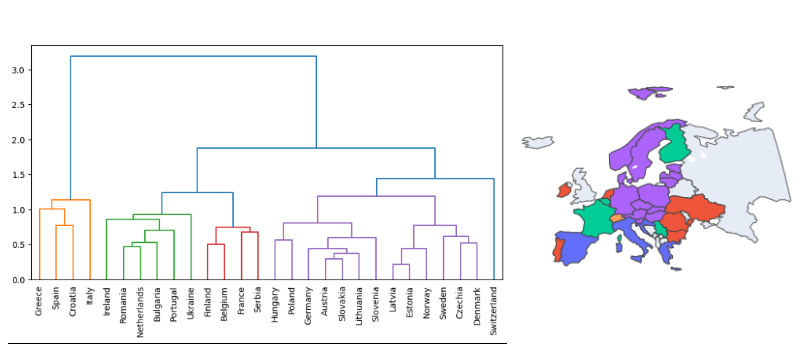}
    \caption{Dendrogram and related choropleth based on clustering done on daily, weekly, and yearly profiles of the countries.}
    \label{fig:4}
\end{figure}

The results of the clustering process are illustrated in Fig.~\ref{fig:4}, where we can observe four basic clusters: (i) Mediterranean countries (cluster 1), (ii) central European countries (cluster 2), (iii) Eastern-European countries (cluster 3), (iv) Scandinavian and Baltic countries (cluster 4). Given these results, that imply certain correlation among the countries of the same cluster, we additionally come up with several geographical and socio-economic facts that can be used to validate the results of the derived grouping by the clustering algorithm:  

\begin{enumerate}
    \item Baltic states (cluster 4): Estonia, Latvia, and Lithuania share historical ties and experiences among the Baltic states, particularly their time under Soviet authority. Since attaining their independence, they have worked to improve their collaboration.
    \item Mediterranean sea (cluster 1): Spain, Greece, Italy, and Croatia share many connections due to their unique geographical position. To develop a more integrated energy market in the Mediterranean region, many of these nations work together on electricity and gas interconnections. They seek to increase cross-border gas and electricity trade to guarantee a steady supply of energy. 
    \item Visegrad Group (cluster 2): Poland, Slovakia, and Hungary have made significant progress at building transnational energy pipelines, electrical connections, and transportation networks.
    \item Benelux (cluster 2): Belgium and the Netherlands share a commitment for free trade as well as a history of economic cooperation, while they are also renowned for their close proximity and highly advanced logistics and transportation networks.
    \item Scandinavia (cluster 4): Sweden, Norway, and Denmark adopt a similar social structure, close proximity to one another, and a comparable cultural history. They work together on a variety of local concerns and have close trading relations (energy policies, resource management, and development of RES, among others).
\end{enumerate}

The aforementioned classifications and current affairs imply that the clustering algorithm has sufficiently distinguished among the observed load patterns of the European countries involved in our study.

\subsection{Selected model} \label{sec:mlp}

Regarding the selected model, we have opted for the MLP for several reasons. Compared to other popular NN architectures, MLPs are relatively simpler, more flexible, and similarly skillful in learning how to map inputs to outputs within a wide range of forecasting applications. Their structure allows to examine them with high precision and they do not require the time-consuming network design search process needed for deeper and more state-of-the-art architectures, therefore allowing for the rapid deployment of the experimental approach and interpretable comparisons. This trait also results in a more straight-forward application of TL techniques, enabling the execution of a wider variety of tests compared to larger, and usually computationally intensive architectures. 

The MLP consists exclusively of perceptrons (or neurons), which are organized in one or more layers. Each perceptron of a layer is fed with the outputs of each perceptron of the previous layer and feeds, in turn, each perceptron of the next layer (forward pass). Equation \ref{eq:00} is used to calculate a given neuron's output. 

\begin{equation}
y = f(\sum_{i=1}^{n} (w_ix_i) + b)
\label{eq:00}
\end{equation}
    
The weighted (with weights $w_i$) sum of the outputs ($x_i$) of all the neurons in the previous layer plus a bias term ($b$) is computed first. Then, the activation function $f$ is applied to this result. The purpose of this function is to enable the model to capture non-linear relationships among the inputs (lags in our case) and outputs (forecasts). Various activation functions have been used in literature, such as the sigmoid, hyperbolic tangent (tanh), and sigmoid linear units (SiLU), as well as the rectified linear unit (ReLU), with the latter being among the most popular (see Eq. \ref{eq:01}).
    
\begin{equation}
f(x) = \frac{1}{1 + e^{-x} }
\label{eq:01}
\end{equation}
    
During the training process of a MLP, the back-propagation algorithm is used. The weights and biases of the MLP are iteratively updated with the use of optimizers to ensure the optimal values for its weights and biases. Among multiple optimizer options (e.g. Gradient Descent, Adagrad, Adadeltra) our choice was the adaptive moment estimation algorithm (ADAM) since it is one of the most widely adopted options in current research. 

\subsection{Transfer learning setup} \label{sec:tl}
We define TL as the extension of ML that uses knowledge previously gained from solving a task, in order to improve performance on a subsequent task ($T$). Given the popular study by \citet{Pan2010ALearning} to define TL, we first need to formalize the following: 

Let $D$ be a domain that includes: (i) a feature space ($X$) and (ii) a marginal probability distribution $P(X)$ for said $X$. Each task $T$ (denoted by $T = \{Y, f(\cdot)\}$) consists of: (i) a label space $Y$ and (ii) a function $f(\cdot)$, which is unobserved and can only be inferred from the training data ($\{x_i, y_i\}\ |\ x_i \in X$ and $y_i \in Y$), and $f(.)$ is used to predict the label $f(x)$ of a new instance $x$. A general-purpose definition of TL is stated as follows:


\textit{\textbf{Definition 1:} "Given a source domain $D_S$ and a learning task $T_S$, a target domain $D_T$ and learning task $T_T$, TL aims to help improve the learning of the target predictive function $f(\cdot)$ in $D_T$ using the knowledge in $D_S$ and $T_S$, where $D_S \neq D_T$ or $T_S \neq T_T$ }."

Aiming to adapt this definition to our TL use case, we can come up with the following statements for each one of our TL experiments:

\begin{itemize}
    \item \textbf{Domains ($D_S/D_T$)}: The feature space is the sequence of historical values used to train the model. Specifically, the length of this sequence directly depends on the optimal look-back window ($l$) of each trained NN. Mathematically, this leads to the following source and target feature spaces respectively: $X_S,X_T \in \mathbb{R}^l$). Therefore, the source and target feature spaces are the same within the TL workflow of a given experiment. In addition, the source and target marginal probability distributions for the predictor variable (future energy demand) are different ($P_S(X) \neq P_T(X)$) within the same experiment, since they depend on the countries included during the training (source training) and fine-tuning (target training) procedures. Given the above, the source and target domains differ within our TL studies ($D_S \neq D_T$).
    \item \textbf{Tasks ($T_S/T_T$):} The label space consists of the range of the possible energy demand forecasts, which is directly dependent on the forecast horizon used. The forecast horizon is constant (24 data points) within our day-ahead forecasting setting, causing source and target label spaces to be the same ($Y_S,Y_T \in \mathbb{R}^{24}$). With regards to the $f(\cdot)$ function, it corresponds to our trained NN, whose parameters (weights and biases) are used for the knowledge transfer procedure. Since $f(\cdot)$ is determined by the provided training data, and each experiment contains a different set of countries, thereby source and target $f(\cdot)$ functions are different ($f_S(\cdot) \neq f_T(\cdot)$). Given the above, the source and target tasks differ within our TL studies ($T_S \neq T_T$).
\end{itemize}

Considering these factors, our TL setups fall within the scope of homogeneous ($Y_S=Y_T$ and $X_S=X_T$) inductive transfer learning \citep{Weiss2016ALearning, Zhuang2021ALearning}, which can be defined as follows:

\textit{\textbf{Definition 2} (Inductive Transfer Learning). "Given a source domain $D_S$ and a learning task $T_S$, a target domain $D_T$ and learning task $T_T$, inductive transfer learning aims to help improve the learning of the target predictive function $f(\cdot)$ in $D_T$ using the knowledge in $D_S$ and $T_S$, where $T_S \neq T_T$}"

Regarding the specifics of TL sub-techniques, within our study, we apply warm-start paired with fine-tuning. Warm-start utilizes previously trained weights or parameters. Specifically, warm-start refers to models being trained on the source domain/task and their parameters being subsequently used as initial values for the target model's parameters (parameter transfer). Note that warm-start applies only an initialization of all model's parameters, in contrast to head replacement which initializes all but the final layers. Warm-start initially involves copying the entire solution from the source task to the target task. This is where fine-tuning comes into play to adjust our source task's solution to the target task. Taking into consideration that: (i) normally NNs' weights and biases are typically initialized using random values, which doesn't offer any initial advantageous/disadvantageous bias for the parameters regarding the target task; and that (ii) the source and target are similar in nature (task, domain, or feature space); the warm-start technique allows to start training the target models from a beneficial ("warmer") position on the loss surface, leading to faster convergence and decreased training time, as the optimal solution has already been significantly approached during the training of the source model. 

\begin{figure}[H]
    \centering
    \includegraphics[width=0.8\textwidth]{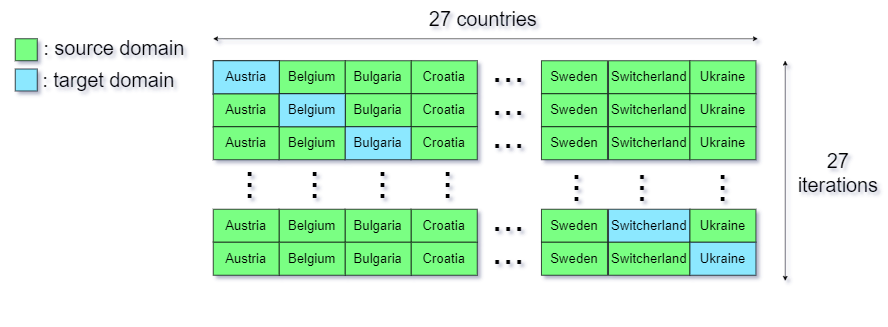}
    \caption{An illustration of the separation process of source and target domains for each experiment in the AbO setup. Each row of cells represents a separate execution of the entire ML pipeline. Source countries on each row are depicted in green, while the corresponding target domain country is depicted in blue. The same approach has been followed in the CbO setup as well, with the exception that this separation takes place internally within the countries of each cluster, excluding those that do not pertain to it.}
    \label{fig:5}
\end{figure}

Aiming to evaluate the added value of TL with or without clustering on the day-ahead national load forecasting of European countries, 3 main setups were implemented to support our comparisons:

\begin{enumerate}
    \item \textbf{Baseline:} We model each country included in the data set individually. In this respect, each country has a unique, individual forecasting model which is trained, optimized, and tested using historical data from said country alone. These models are used as a baseline to evaluate the potential improvements of the two TL setups described below. 
    \item \textbf{All-but-One (AbO):} A TL setup according to which, given a certain country, a model is first pre-trained on the data of all the other countries included in the data set and then fine-tuned using the data of the selected country. As shown in \ref{fig:5}, our data set contains data from 27 different countries. Therefore, according to this setup, we perform experiments equal to the number of countries, where each time a different country is set as the target domain, while the rest 27 are set as the source domain. A model ($f_S$) is trained in the source domain and its parameters are used (via the warm-start technique) to develop a new model ($f_S$) in order to generate forecasts in the target domain.  
    \item \textbf{Cluster-but-One (CbO)}: As stated in section \ref{sec:2.3}, countries with a similar geographical, climatic, and socioeconomic characteristics may also share similarities in their electricity demand. In this context, this TL setup is developed to examine the performance of TL between countries pertaining to the same cluster. The approach is similar to the AbO setup, with the exception that in a given country's experiment, the source domain comprises countries that belong in the same cluster, rather than the entire data set.
\end{enumerate}

For the purpose of completeness and objective assessment, we also introduce a seasonal naive model of weekly seasonality, namely seasonal naive with a look-back window of 168 values \textbf{sNaive(168)}. This naive approach uses the observed values of the same day of the previous week as day-ahead forecasts for each of the countries and is expected to produce moderate results (daily and weekly seasonalities are incorporated), thus supporting the objective benchmarking of the TL and baseline approaches \citep{Pelekis2023ADrivers}. 


\subsection{Model training and validation pipeline} \label{sec:3.4}
In section \ref{sec:2.2} we described the MLP architecture selected for our models. As mentioned, during the model training and validation procedure, all models have been optimized to minimize the L2 loss, using the ADAM optimizer. Additionally, early stopping was applied with a patience of 10 epochs. 

Our data set contains data from 27 different countries. Therefore,  depending on the setup and target domain, each model might be subject to different (but mostly overlapping) data points. In this context, the training process can be summarized as follows: 
\begin{itemize}
    \item each model is trained on a train set spanning from 2015 to 2019, that is $N_s$ - $N_t$ unique data points, where $N_s$ is the number of observations pertaining to the countries of the source domain and $N_t$ is the number of observations belonging to the target country. Note that $N_s$ can vary depending on (i) the historical data availability of each country and (ii) the TL setup. For example, the AbO setup results to much larger training data sets than CbO. The $N_t$ variable is affected by the fact that not all countries' observations date back to 2015, which leads to variant population size among target domains. This is not the case for validation and test sets, as all time series have available observations for the years 2020 and 2021.  
    \item each model is optimized on the validation set (year 2020: 239,018 unique data points) to identify appropriate hyperparameter values.
    \item each model is evaluated on the remaining, and previously unseen test set (year 2021: 243,541 unique data points) without retraining the model on the full data set (union of the training and validation sets).
\end{itemize}

Note that the aforementioned data set splitting process allowed the inclusion of complete calendar years within each data set, thus maintaining the symmetry of seasonal patterns.

For each experiment, the hyperparameters of the MLPs were tuned using a multi-dimensional grid of hyperparameter values and more specifically the tree-structured Parzen estimator (TPE) optimization approach, as described by \citet{Bergstra2011AlgorithmsOptimization} and implemented in Python (version 3.8) programming language in Optuna optimization library (version 2.10) \citep{Akiba2019Optuna:Framework}, similar to \citet{Pelekis2023ADrivers}. Additionally, the successive halving algorithm -- a combination of random search and early stopping -- was internally applied as a pruning criterion for TPE trials. A crucial hyperparameter that is common among all architectures, is the look-back window ($l$), namely the number of historical time series values (lags) that the model looks back at when being trained to produce forecasts. The $l$ directly determines the input layer size of the trained neural network. The specific details of the optimization process for each architecture are presented in Tab. \ref{tab:1}. A total of 100 trials were executed for the optimization of each NN architecture. 

Following the identification of the best model architecture in terms of hyperparameter values, 20 separate networks sharing the same hyperparameter values but different pseudo-random initialization of neural weights were trained for each NN architecture. Then, an ensemble of these models was used to produce the final forecasts. To improve the robustness of the results, the final forecast of our ensemble model was derived as the average of the predictions of all models. The ensemble method was performed using Ensemble-PyTorch (version 0.1.9) \citep{EnsemblePyTorchEnsembleDocumentation}, a unified framework for ensembling in PyTorch, with a user-friendly API for ensemble training/evaluation, as well as high efficiency in training with parallelism.  

During the TL step, hyperparameter tuning was bypassed, as the hyperparameters and trainable parameters of the given source model were passed, as were, for the initialization of the target model (warm-start technique). The re-training of the target model followed, being fine-tuned until its validation loss plateaued out (early stopping).

With respect to computing resources, the training, validation, and testing of the models was executed on an Ubuntu 22.04 virtual machine with an NVIDIA Tesla V100 GPU, 32 CPU cores, and 64GB of RAM.

\begin{table}[H]
\centering
\caption{The hyperparameters optimized per NN architecture and the respective search spaces.}\label{tab:1}
\begin{tabular}{@{}lll@{}}
    \toprule
    \multicolumn{1}{l}{\textbf{Arguments}} & \multicolumn{1}{l}{\textbf{Value Range}} & \multicolumn{1}{l}{\textbf{Type}} \\
    \midrule
    \centering
    number of layers &  \{2, ... ,6\} & Discrete\\ 
    layer sizes & \{128, 256, 512, 1024, 2048\} & Discrete \\ 
    $l$ &  \{168, 336, 504, 672\} & Discrete \\ 
    forecast horizon & 24 & Fixed \\ 
    learning rate &  \{$1e^{-5}$, ... , $1e^{-4}$\} & Continuous \\ 
    batch size & \{256, 512, 1024\} & Discrete \\  
    \bottomrule
\end{tabular} \\
\end{table}

\subsection{Model evaluation} \label{sec:3.5}
Various forecasting performance measures are common in the literature for evaluating forecasting accuracy, such as the mean absolute error (MAE), the mean squared error (MSE), the mean absolute percentage error (MAPE), the root mean square error (RMSE), the symmetric mean absolute percentage error (sMAPE), and the mean absolute scaled error (MASE) \citet{Hyndman2006}. Within the present study, MAPE was selected as it is a widely accepted choice in STLF applications and results to interpretable measurements. Consequently, MAPE served as the objective function during the hyperparameter optimization process and as the main evaluation measure. MAPE is defined as follows:

\begin{equation}
    MAPE=\frac{1}{m} \sum_{i=1}^{m}\left|\frac{Y_{t}-F_{t}}{Y_{t}}\right| \cdot 100(\%),
    \label{eq:1}
\end{equation}

\noindent where $m$ represents the number of samples, while $Y_t$ and $F_t$ stand for the actual values and the forecasts at time $t$, respectively. 

\section{Results and Discussion}\label{sec:4}

\begin{figure}[htb]
    \centering
    \includegraphics[width=0.8\textwidth]{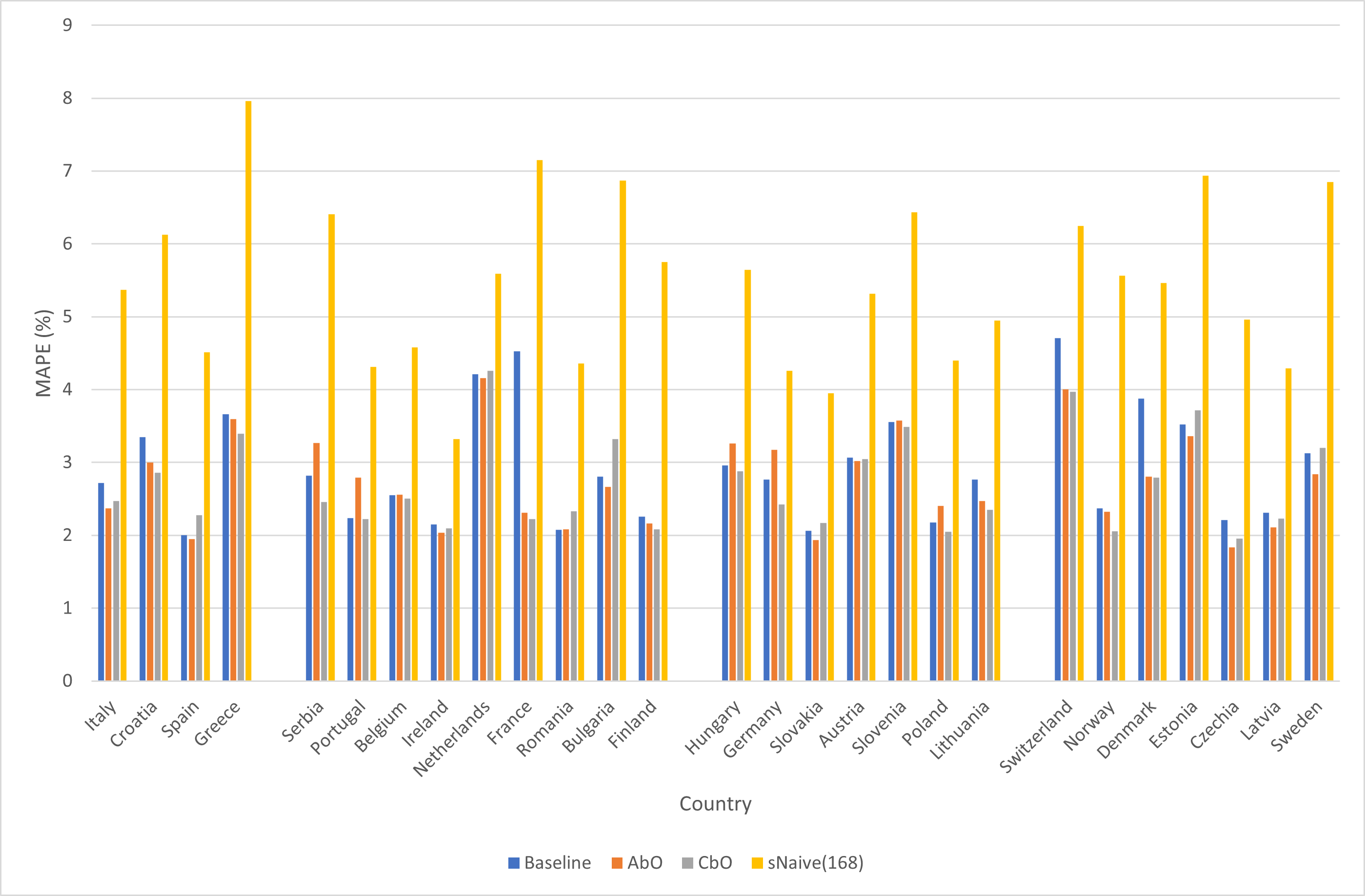}
    \caption{Side-by-side barplots depicting the MAPE(\%) for each TL setup and baseline for each country}
    \label{fig:6}
\end{figure}

\begin{figure}[htb]
    \centering
    \includegraphics[width=0.8\textwidth]{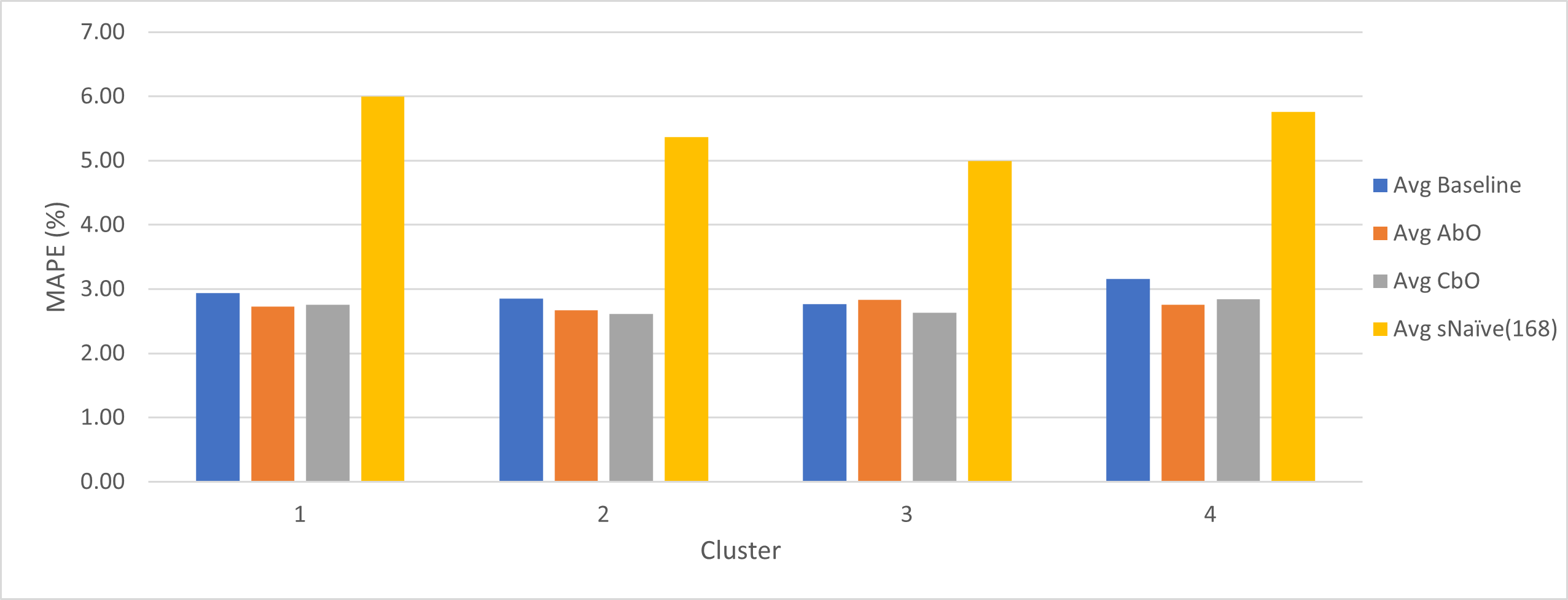}
    \caption{Side-by-side barplots depicting the average MAPE(\%) for each TL setup and the baseline for each cluster}
    \label{fig:7}
\end{figure}

Proceeding to the results of our experiments, the final performance of each architecture is displayed in Fig. \ref{fig:6} and \ref{fig:7} (derived from Tab. \ref{tab:3} and \ref{tab:4}, respectively), taking into account the ensembling modeling technique that followed the hyperparameter optimization process. More details on the optimal hyperparameters and training times of the best performing models can be found in Tab. \ref{tab:2} of the appendix \ref{app:a}. As an initial observation, sNaive(168) is heavily outperformed at all cases by all NN models (AbO, CbO, baseline). This is but an expected outcome, since naive models do not entail any kind of learning process; they only repeat past values of the time series. In fact, this model was selected as a hard baseline, helping to ensure the validity of our setups rather than to actually benchmark them. 

Focusing on the effectiveness of TL, Fig. \ref{fig:7} demonstrates that within each cluster, at least one of 2 TL setups, achieve, on average, a lower forecasting error than the baseline. Therefore, it becomes apparent that TL, as a whole, outperforms at all cases the traditional NN approach that is represented by our baseline model. Note here that the TL approaches, in total, reduce said forecasting error, on average, by 0.34\% with respect to MAPE, as shown in Tab.~\ref{tab:3}. Additionally, note that TL enables the advantageous initialization of target models from their respective source models, leading to much faster convergence and therefore a significant decrease in the target pipeline's (fine-tuning) execution compared to the respective baseline model execution. Said decrease in execution time ranges from 12.09\% up to 66.87\% (average 47\%) depending on the experiment (target country). More details regarding the exact training times per experiment can be sought in in Tab. \ref{tab:2} of the appendix \ref{app:a}.

Considering the analytical results of Tab. \ref{tab:3} and Fig. \ref{fig:6}, it is notable that the AbO setup results to an average error reduction of 0.16\% compared to the baseline in terms of absolute MAPE percentage. Therefore, this setup, outperforms the baseline on average, even if this is not the case for all the countries in our data set. In the same direction, CbO leads to a higher average error reduction of 0.24\% compared to the baseline, therefore confirming the value added by the clustering scheme proposed within our study. 

\begin{table}[htb]
\centering
\caption{Accuracy (MAPE) of the examined forecasting approaches (baseline, AbO, CbO) alongside the naive model sNaive(168). The last row lists the average improvement of each model compared to the baseline model. The last column contains the MAPE of the best performing model alongside its name. Note here that AbO or CbO are the only alternating values demonstrating that there is always a TL setup that outperforms the rest. In this context, the last cell of the table corresponds to the average improvement of TL (0.34 \%), in general, compared to the NN baseline model.}
\begin{tabular}{@{}ll|lllll@{}}
\toprule
{\textbf{Country}} & {\textbf{Cluster No.}} & \multicolumn{5}{c}{\textbf{MAPE \%}} \\
~ & ~ & Baseline & AbO & CbO & sNaive(168) & Best model\\
\midrule
\textbf{Italy} &  1 & 2.72 & \textbf{2.37}  & 2.47 & 5.37 & 2.37 (AbO)\\
\textbf{Croatia} & 1 & 3.35 & 3.00  & \textbf{2.86} & 6.13 & 2.86 (CbO)\\
\textbf{Spain} & 1 & 2.00 & \textbf{1.95} & 2.28 & 4.51 & 1.95 (AbO)\\
\textbf{Greece} & 1 & 3.66 & 3.60  & \textbf{3.39} & 7.97 & 3.39 (CbO)\\
\textbf{Serbia} & 2 & 2.82 & 3.27  & \textbf{2.46} & 6.41 & 2.46 (CbO)\\
\textbf{Portugal} & 2 & 2.24 & 2.79  & \textbf{2.23} & 4.31 & 2.23 (CbO)\\
\textbf{Belgium} & 2 & 2.55 & 2.56  & \textbf{2.50} & 4.58 & 2.50 (CbO)\\
\textbf{Ireland} & 2 & 2.15 & \textbf{2.03}  & 2.09 & 3.32 & 2.03 (AbO)\\
\textbf{Netherlands} & 2 & 4.21  & \textbf{4.16} & 4.26 & 5.59 & 4.16 (AbO)\\
\textbf{France}  & 2 & 4.53 & 2.31  & \textbf{2.22} & 7.15 & 2.22 (CbO)\\
\textbf{Romania}  & 2 & 2.54 & \textbf{2.09}  & 2.33 & 4.36 & 2.09 (AbO) \\
\textbf{Bulgaria} & 2 & 2.80 & \textbf{2.67}  & 3.32 & 6.87 & 2.67 (AbO)\\
\textbf{Finland}  & 2 & 2.26  & 2.16  & \textbf{2.08} & 5.75 & 2.08 (CbO)\\
\textbf{Hungary} & 3 & 2.96 & 3.26  & \textbf{2.88} & 5.64 & 2.88 (CbO) \\
\textbf{Germany} & 3 & 2.76 & 3.17  & \textbf{2.42}  & 4.26 & 2.42 (CbO)\\
\textbf{Slovakia} & 3 & 2.06 & \textbf{1.94}  & 2.17 & 3.95 & 1.94 (AbO) \\
\textbf{Austria} & 3 & 3.07  & \textbf{3.02}  & 3.04 & 5.32 & 3.02 (AbO)\\
\textbf{Slovenia} & 3 & 3.56 & 3.58  & \textbf{3.49}  & 6.44 & 3.49 (CbO)\\
\textbf{Poland} & 3 & 2.18 & 2.40  & \textbf{2.05}  & 4.39 & 2.05(CbO)\\
\textbf{Lithuania} & 3 & 2.77 & 2.47  & \textbf{2.35} & 4.95 & 2.35 (CbO)\\
\textbf{Switzerland} & 4 & 4.71 & 4.01  & \textbf{3.97} & 6.25 & 3.97 (CbO)\\
\textbf{Norway} & 4 & 2.37 & 2.33  & \textbf{2.06} & 5.56 & 2.33 (CbO)\\
\textbf{Denmark} & 4 & 3.87 & 2.80  & \textbf{2.79} & 5.46 & 2.79 (CbO) \\
\textbf{Estonia} & 4 & 3.52 & \textbf{3.36}  & 3.72 & 6.94 & 3.36 (AbO) \\
\textbf{Czechia} & 4 & 2.21  & \textbf{1.83}  & 1.96  & 4.97 & 1.83 (AbO)\\
\textbf{Latvia} & 4 & 2.31 & \textbf{2.11}  & 2.23 & 4.29 & 2.11(AbO)\\
\textbf{Sweden} & 4 & 3.13 & \textbf{2.84} & 3.19 & 6.85 & 2.84 (AbO)\\
\midrule
\textbf{Average Improvement} &  & - & 0.18 & 0.24 & -1.8 & 0.34 (TL) \\
\bottomrule
\end{tabular}
\label{tab:3}
\end{table}

\begin{table}[htb]
\captionsetup{font=normalsize}
\centering
\caption{Average MAPE(\%) per cluster and training setup}
\begin{tabular}{@{}cccccc@{}}
\toprule
\multicolumn{1}{l}{\textbf{Cluster no.}} &
  \multicolumn{1}{l}{\textbf{Baseline}} &
  \multicolumn{1}{l}{\textbf{AbO}} &
  \multicolumn{1}{l}{\textbf{CbO}} &
  \multicolumn{1}{l}{\textbf{sNaive(168)}} & 
  \multicolumn{1}{l}{\textbf{Best setup}} \\
\midrule
\textbf{1} & 2.93 & \textbf{2.73} & 2.75 & 5.99 & AbO \\
\textbf{2} & 2.85 & 2.67 & \textbf{2.61} & 5.37 & CbO\\
\textbf{3} & 2.77 & 2.83 & \textbf{2.63} & 4.99 & CbO\\
\textbf{4} & 3.16 & \textbf{2.75} & 2.85 & 5.75 & AbO\\
\bottomrule
\end{tabular}
\label{tab:4}
\end{table}

\subsection{A note on the best-performing TL setup}
 
With respect to CbO, its average improvement (0.24\%) compared to the baseline is higher than AbO (0.18\%) and therefore has already been considered as the best performing TL setup. To further reinforce this statement, it can be noted that CbO also holds the majority of countries where it achieved better results than its TL competitor (AbO). More specifically, among the 27 European countries that were examined in total, 12 ($44.5$\%) of them demonstrate better results with the AbO setup compared to CbO, while the opposite being true for the other 15 ($55.5$\%). This outcome can be attributed to the stratified training data selection proposed by our clustering methodology. In this direction, by providing a specific, highly correlated group of countries as its training set, the target model can take advantage of the similarities among countries and transfer the required knowledge in an optimal fashion. Despite the fact that CbO proved to be quite effective in most of the countries, the shortcomings of said method cannot be neglected. Countries like Spain, Romania, and Bulgaria exhibit reduced performance with CbO compared to the respective baseline MLP. 

Focusing on Tab.~\ref{tab:4} and Fig.~\ref{fig:7}, an interesting observation lies in the fact that cluster 3 is the only case where AbO is significantly outperformed, even by the baseline model. This finding further validates the superiority of the clustering scheme, confirming that CbO is the most consistent TL setup. Regarding the poor performance of AbO in cluster 3, it should be noted however that its forecast error is not always worse than the baseline NN, being higher mainly in four (Germany, Poland, and Hungary) of the seven countries included in the cluster. 


\subsection{A note on selecting the optimal TL setup based on the target country}
Another interesting discussion topic relates to the selection between AbO and CbO depending on the target country of interest. Our results suggest that said selection should be evaluated individually per use case, and therefore specific caution is advised during the selection process. However, a good indicator for clustering could, by definition, be the degree of similarity within the energy demand patterns observed among the countries of each cluster. This can be further supported by the significant overlap between countries, where clustering was the most effective for countries that indeed have several interconnections with each other. In this direction, several interesting insights can be extracted from the official ENTSO-E transmission system map 
\citep{GridMap}, as follows:

\begin{enumerate}
    \item \textbf{Cluster 1:} Croatia, Italy and Greece have made a lot of progress developing energy interconnections between them. There is an established 380-400 kV Transmission Line between Redipuglia (Italy) and Melina (Croatia) as well as a 400KV HVDC link between Galatina (Italy) and Arachthos (Greece)
    \item \textbf{Cluster 2:} Belgium and the Netherlands are closely energy related and have created transmission lines between  and Van Eyck–Maasbracht interconnection. Additionally, France and Belgium have also developed cross-border interconnections  
    between Mastaing-Avelgem and Avelin-Avelgem  
    \item \textbf{Cluster 3:} Austria is connected with several interconnections with Hungrary, Germany, Slovenia and Slovakia (e.g Kainarkdal-Cirkovce, Silz-Oberbrumm etc). Poland and Lithuania are also connected through various electricity interconnections, including the LitPol Link: a high-voltage direct current (HVDC) interconnection that allows for electricity exchange between the two countries.
    \item \textbf{Cluster 4:} Sweden has a strong energy connection with Norway, primarily through hydroelectric power generation
    as well as Denmark through electricity transmission lines and substations. Estonia and Latvia are also connected through electricity interconnections (e.g Valmiera-Trisguliina), promoting regional energy cooperation and grid stability.
\end{enumerate}
An interesting observation lies in cluster 1. Croatia, Italy, and Greece have developed energy interconnections with each other, while, on the contrary, Spain has not -- at least as the time of writing this paper. In this context, Spain is the only country in its cluster where CbO was outperformed by the NN baseline (and AbO by extension). Despite the appeal of this theory, there are a few exceptions, such as the case of Sweden which did not respond as expected to the CbO approach.

\section{Conclusions and future research}\label{sec:6}

This study conducted a comparative analysis of TL approaches used for day-ahead national net aggregated electricity load forecasting. A feed-forward NN architecture was selected to implement the baseline forecasting models, followed by proper hyperparameter tuning and ensembling. The warm-start technique, accompanied by fine-tuning, was employed as a widely used and streamlined TL technique. Within our initial approach (AbO), we examined the value of straight-forward TL. We developed a separate model for each country of the data set, aiming at forecasting its energy load (target domain). Each model draw its initial parameters' values from another unique model, pre-trained using the data from the rest of the countries included in the data set (source domain). Moreover, motivated by an exploratory data analysis, which revealed a certain correlation among the energy load patterns of different countries -- due to similarities in their geographical and socioeconomic conditions -- we investigated if the identified correlations could be exploited to further boost forecasting performance. In this direction, we developed a hierarchical clustering algorithm to derive clusters of countries and then used them to form the proper source domains for the TL pipeline by pre-training the respective global model using only the time series of the countries pertaining to the same cluster with the target country (\textit{Cb0} setup). 

Following our experiments, the results suggest that at least one of the examined TL approaches (AbO or CbO) outperforms the baseline NN model. Specifically, the clustering-based TL pipeline (CbO) outperforms the rest of the NN models, improving the average MAPE by 0.18\% when compared to AbO and 0.24\% when compared to the baseline. Despite its benefits, CbO often exhibited worse performance, especially for countries with outlying socioeconomic profiles and/or load patterns and magnitudes. In such cases, the AbO model is recommended to counter the negative bias that clustering entails. 

The methodology and results of our study can be used by researchers and forecasting practitioners, alongside certain stakeholders of the energy sector, such as transmission and distribution system operators, to develop unified TL-based models for previously unseen national load time series. Said models combine both increased accuracy and an average training acceleration of 47\% compared to their respective, conventional NN approach. 

Regarding future perspectives, it would be of high interest to explore the clusters and investigate their pertaining countries more thoroughly with respect to political and socioeconomic factors, such as energy policies, energy providers, and cross-country grid interconnections. Such an analysis could allow the selection of more appropriate clustering algorithms or even the manual classification of countries aiming to an optimized CbO performance. Furthermore, a possible future next step of this research could involve the experimentation and TL approach with more complex NN architectures as baseline models, such as LSTM, CNN, or transformers. With respect to the specifics of TL, the use of additional TL techniques (e.g freezing, head replacement/feature extraction) is also highly recommended in an attempt optimize the TL pipeline, leading to further increase of both the accuracy and computational efficiency of the models. With respect to available data sets, further performance increase could be potentially achieved by (i) enriching the utilized electricity demand data set with non-European countries and (ii) incorporating weather forecasts or energy price fluctuations within the training procedure. Finally, future research could involve the comparison of global forecasting models with the existing TL models (AbO, CbO), alongside their zero-shot versions, i.e. approaches that omit the fine-tuning stage. Such a comparison would lead to interesting results regarding the cases where (a) target data sets are (global) or are not (TL) available during the model training process and (b) fine-tuning is hard to conduct after handing the pre-trained model (zero-shot TL case).
    

\section*{Acknowledgment}
This work has been funded by the European Union’s Horizon 2020 research and innovation program under the I-NERGY project, grant agreement No. 101016508. Additionally, the HPC resources utilized for training and optimizing the required ML models in this study have been provided by the EGI-ACE project, which also receives funding from the European Union’s Horizon 2020 research and innovation program under grant agreement No. 101017567.

\bibliographystyle{unsrtnat}

\bibliography{references}


\appendix
\section{Tables} \label{app:a}
\renewcommand\thetable{\thesection.\arabic{table}}
\setcounter{table}{0} 


\begin{table}[H]
\centering
\caption{Training time and hyperparameters of the best-performing setups in terms of MAPE. Each line of table represents a country, containing (a) the name of its best performing setup (b) the training time of the target model (fine-tuning process) (c) the training time of the baseline model (d) hyperparameters of the best model.} \label{tab:2}
\begin{adjustbox}{width=\textwidth, totalheight=\textheight, keepaspectratio}
\small
\begin{tabular}{@{}ll|lll|lllll@{}}
\toprule
\textbf{Country} & \textbf{Best performing setup} & \multicolumn{3}{c|}{\textbf{Pipeline duration (minutes)}} & \multicolumn{5}{c}{\textbf{Best hyperparameters}}     \\
~ & ~ & source & target & baseline & lookback window & learning rate & layer number & layer sizes & batch size \\
\midrule
Italy       &  AbO & 402    & 20  & 51 & 168 & 0.000206 & 4 & \{1024,256,2048,1024\}   & 1024 \\
Croatia     &  CbO & 90   & 17  & 40 & 168 & 0.000162 & 3 & \{256,256,1024\}         & 256  \\
Spain       &  AbO & 402  & 16  & 50 & 168 & 0.000224 & 2 & \{256,2048\}             & 512  \\
Greece      &  CbO & 72   & 24  & 58 & 504 & 0.000313 & 5 & \{512,256,512,128,2048\} & 1024 \\
Serbia      &  CbO & 126   & 21  & 38 & 168 & 0.000617 & 3 & \{2048,1024,1024\}       & 256  \\
Portugal    &  CbO & 168  & 29    & 44 & 672 & 0.000541 & 4 & \{1024,2048,512,1024\}   & 512  \\
Belgium     &  CbO & 114  & 21    & 47   & 168 & 0.000839 & 2 & \{128,512\}              & 256  \\
Ireland     &  AbO & 546  & 29  & 33 & 336 & 0.000116 & 4 & \{512,256,512,2048\}     & 256  \\
Netherlands & AbO  & 492  & 23  & 55 & 168 & 0.000511 & 3 & \{1024,2048,128\}        & 256  \\
France      & CbO  & 138   & 25  & 46 & 168 & 0.000428 & 3 & \{512,2048,2048\}        & 256  \\
Romania     & AbO  & 534  & 25  & 46 & 336 & 0.000213 & 3 & \{512,256,256\}          & 256  \\
Bulgaria    & AbO  & 1134    & 32  & 47 & 336 & 0.000169 & 4 & \{2048,256,1024,256\}    & 256  \\
Finland     & CbO  & 156  & 23  & 46 & 168 & 0.000307 & 3 & \{512,1024,256\}         & 256  \\
Hungary     & CbO  & 132   & 26  & 46 & 168 & 0.000928 & 4 & \{128,512,512,128\}      & 256  \\
Germany     & CbO & 156   & 14   & 31 & 504 & 0.000810 & 4 & \{128,256,128,2048\}     & 256  \\
Slovakia    & AbO & 600   & 26  & 66   & 168 & 0.000130 & 3 & \{2048,256,512\}         & 256  \\
Austria     & AbO & 348   & 19  & 49   & 168 & 0.000644 & 4 & \{2048,128,512,256\}     & 512  \\
Slovenia    & CbO & 168   & 29    & 36 & 168 & 0.000385 & 5 & \{512,256,128,128,128\}  & 256  \\
Poland      & CbO & 150   & 25  & 54 & 504 & 0.000368 & 4 & \{128,2048,1024,256\}    & 512  \\
Lithuania   & CbO & 174   & 25  & 32 & 504 & 0.000240 & 2 & \{1024,1024\}            & 256  \\
Switzerland & CbO & 162   & 28  & 54 & 336 & 0.000136 & 5 & \{1024,256,2048,512,512\}& 256  \\
Norway      & CbO & 150   & 22  & 51 & 168 & 0.000188 & 2 & \{128,1024\}             & 256  \\
Denmark     & CbO & 126     & 20  & 44 & 168 & 0.000140 & 3 & \{2048,1024,512\}        & 512  \\
Estonia     & AbO & 546   & 25  & 58   & 336 & 0.000117 & 4 & \{512,1024,512,2048\}    & 256  \\
Czechia     & AbO & 498   & 24  & 44 & 336 & 0.000165 & 2 & \{1024,2048\}            & 256  \\
Latvia      & AbO & 1584   & 40  & 46 & 168 & 0.000124 & 4 & \{1024,256,2048,512\}    & 256  \\
Sweden      & AbO & 570   & 30  & 60   & 168 & 0.000278 & 4 & \{256,512,256,128\}      & 256 \\
\bottomrule
\end{tabular}
\end{adjustbox}
\end{table}

\end{document}